\documentclass{article}

 \usepackage[main, final]{neurips_2026}

\usepackage[utf8]{inputenc}
\usepackage[T1]{fontenc}
\usepackage{hyperref}
\usepackage{url}
\usepackage{booktabs}
\usepackage{amsfonts}
\usepackage{nicefrac}
\usepackage{microtype}
\usepackage{xcolor}

\usepackage[table]{xcolor}
\usepackage{multirow}
\usepackage{graphicx}
\usepackage{tikz}
\usepackage{amsmath}
\usepackage{amsthm}
\usepackage{caption}
\theoremstyle{plain}
\newtheorem{theorem}{Theorem}[section]
\newtheorem{proposition}[theorem]{Proposition}
\newtheorem{lemma}[theorem]{Lemma}
\newtheorem{corollary}[theorem]{Corollary}
\theoremstyle{definition}

\theoremstyle{remark}

\newcommand{\name}{DnNP}
\newcommand{\obin}{x}
\newcommand{\nfin}{{x^*}}

\title{\name{}: Denoising Input Uncertainty  \\in  Neural Processes}

\author{%
  Fatemeh Tohidian$^{1}$ \quad
  Chengzhi Shi$^{1}$ \quad
  Soomi Lee$^{1}$ \quad
  Matthew Elia$^{2}$ \\
  \textbf{Amy Mueller}$^{2,3}$ \quad
  \textbf{Stratis Ioannidis}$^{1}$ \quad
  \textbf{Jennifer Dy}$^{1}$ \\[4pt]
  \tiny{
 $^{1}$Electrical \& Computer Engineering \quad
  $^{2}$Civil \& Environmental Engineering \quad
  $^{3}$Marine \& Environmental Sciences
  } \\
  \tiny{Northeastern University, Boston, MA, USA}
}

\begin{document}

\maketitle

\begin{abstract}
  Neural Processes (NPs) are flexible meta-learning models capable of uncertainty quantification across diverse applications. However, the standard NP framework assumes access to noise-free inputs. This is unrealistic in many practical settings, where sensor noise or measurement errors are unavoidable. This discrepancy between model assumptions and real-world data can significantly degrade prediction quality and uncertainty estimates. We introduce Denoising Neural Processes (\name{}s), a principled extension of the NP framework that explicitly models input uncertainty.  We train \name{}s via variational inference. \name{}s jointly learn to denoise corrupted inputs and perform meta-learning, making them robust to noisy training data. We demonstrate \name{}s' effectiveness on both synthetic and real-world datasets, showing consistent improvements in prediction and uncertainty quantification compared to standard NPs. On PM$_{2.5}$ spatial 
interpolation from real-world air quality sensors, \name{} reduces RMSE 
by up to 30\% and NLL by up to 73\% relative to standard NPs. We also provide theoretical grounding: we show that the Bayes-optimal predictor under noisy inputs belongs to the \name{} model class, and that plug-in models with fixed observation variance incur an irreducible KL divergence gap that \name{} provably avoids. Code is available at \url{https://github.com/iSUPER-ML/DnNP}.
\end{abstract}

\section{Introduction}
\label{sec:intro}

Meta-learning enables models to quickly adapt to new tasks by learning from many related tasks \citep{meta_learning_survey, model_ag_meta_learning}. Neural Processes (NPs) are flexible meta-learning models that provide uncertainty quantification and have shown strong performance across diverse applications \citep{np_survey}. However, NPs assume that inputs are observed without noise. In practice, inputs are often collected from sensors with measurement uncertainty \citep{metro_sensor_noise}. This mismatch between model assumptions and real-world data can degrade prediction quality.

Classic NPs ~\citep{cnp, np, convcnp} do not model input noise. Inputs are treated as deterministic observations, and all uncertainty is captured through the latent variable and the output likelihood~\citep{np}. Although the problem of handling label noise for NPs has been studied ~\citep{noisy_y_np}, training NPs with inputs contaminated by noise remains unaddressed. This problem has been studied for Gaussian Processes ~\citep{noisy_gp}, another method that outputs distributions over functions in a principled manner ~\citep{gp}. 
Moreover, handling noisy inputs in NPs is challenging for several reasons. Noisy inputs corrupt the learned relationships between features and outputs. This leads to both degraded predictions and miscalibrated uncertainty estimates that fail to account for measurement noise.

We propose Denoising Neural Processes (\name{}), which jointly learn to denoise inputs and make predictions with uncertainty quantification. We derive a principled training objective using variational inference that explicitly accounts for input uncertainty, which is applicable to other NP variants that follow the standard NP framework. We evaluate \name{} across multiple scenarios including synthetic datasets and real-world air quality data, demonstrating consistent improvements. To summarize,
our contributions are as follows:

\begin{itemize}
    \item We propose \name{}, a Neural Process model that explicitly 
    handles noisy input coordinates through a variational denoising 
    mechanism, extending the NP family to a practically important 
    but previously unaddressed setting.

    \item We provide theoretical guarantees showing that the Bayes-optimal predictor under noisy coordinates belongs to the \name{} 
    model class, and that plug-in models with fixed observation 
    variance incur an irreducible KL divergence gap that \name{} 
    provably avoids.

    \item We demonstrate the effectiveness of \name{} on real-world 
    PM$_{2.5}$ spatial interpolation on an Air Quality dataset, achieving up 
    to 30\% reduction in RMSE and 73\% reduction in NLL  over standard 
    NPs across multiple feature configurations (Table~\ref{tab:spatial_interpolation}).
\end{itemize}

\section{Related Work}

    Neural Processes (NPs) have shown strong performance for function estimation across various applications, including weather forecasting ~\citep{np_bsa, np_weather}, spatiotemporal prediction ~\citep{rnp,ranp, nspp}, and medical image analysis ~\citep{fac_np}. NPs leverage the fast and scalable nature of neural networks, while providing a distribution over functions like Gaussian Processes (GPs). The first model of the Neural Processes family, Conditional Neural Processes (CNPs) ~\citep{cnp}, was inspired by the flexibility of GPs while remaining scalable to large and complex datasets. Latent Neural Processes (LNPs) ~\citep{np} introduce a global latent variable that enables sampling functions which better match the underlying distribution by capturing global structure and uncertainty, addressing the independent prediction issue of CNPs, whereby predictions at different target points are made independently and thus lack coherence across the output space. 
    
    Attentive Neural Processes (ANPs) ~\citep{anp} leverage attention to allow each target prediction to selectively attend to the most relevant context points, in contrast to prior work that treats all context points equally through mean pooling. By explicitly modeling translation equivariance, a suitable structural assumption when the underlying process is stationary, Convolutional Conditional Neural Processes (ConvCNPs) ~\citep{convcnp} broaden the applicability of the Neural Processes family. Transformer Neural Processes (TNPs) ~\citep{tnp} adopt a transformer architecture, recasting function approximation as a sequence modeling problem. Distance-Informed Neural Processes (DNPs) ~\citep{dnp} enrich the latent space by adding a local latent variable. Functional Neural Processes (FNPs) ~\citep{fnp} learn graph-based dependencies between data points rather than relying solely on global latent variables.
Despite these advances, all the mentioned variants assume that the model inputs are noise-free, which causes overconfident and inaccurate predictions at test time in real-world settings where inputs are corrupted by noise from various sources.  

Robust Neural Processes ~\citep{noisy_y_np} introduce a training strategy to overcome label noise through two modifications to the standard NP loss function: an additional variance penalty term that discourages the model from inflating pointwise variance, instead encouraging uncertainty to be expressed at the global function level, and restricting the reconstruction loss to target points only to prevent memorization of noisy context values. In contrast, DnNP differs not only in the type of noise addressed, inputs  $x$ rather than labels $y$, but also in the nature of the contribution itself: while Robust NPs propose a training strategy applied to existing architectures, DnNP introduces changes to both the generative model and the architecture.

Gaussian Processes (GPs) have motivated most design choices in the NP family. Gaussian Process Training with Input Noise ~\citep{noisy_gp} addresses the problem of noisy inputs  by applying a first-order Taylor expansion around the noisy observations as a proxy for the latent noise-free inputs. While an exact treatment would require maintaining a distribution over Taylor expansions, whose first and second moments are analytically tractable but computationally prohibitive, the authors instead differentiate only the GP mean function. This yields a linear model for the input noise, adding a term to the likelihood proportional to the squared posterior mean gradient. \name{} instead treats noise-free inputs as latent variables and learns a variational posterior over them within the NP meta-learning framework, enabling amortized inference across tasks.

\section{Background}
\label{sec:background}

Let $\mathcal{X}\subseteq\mathbb{R}^d$ and $\mathcal{Y}=\mathbb{R}$.
A regression \emph{episode} (task) $\mathcal{D}=\{(\obin_i,y_i)\}_{i=1}^n$ consists of input--output pairs,  where $\obin_i \in \mathcal{X}$ 
and $y_i \in \mathcal{Y}$, \(i\in [n]\equiv\{1,\dots,n\}\), are generated from an episode-specific 
latent function $g:\mathcal{X} \rightarrow \mathcal{Y}$. 
We partition indices $[n]$ into a \emph{context} set $C$ and \emph{target} set $T$, where 
$C\cup T=[n]$.
For $S\subseteq [n]$, denote by $\mathcal{D}_S=\{(\obin_i,y_i)\}_{i\in S}$, $\obin_S=\{\obin_i\}_{i\in S} $ and $y_S =\{y_i\}_{i\in S}$,  the corresponding restricted datasets, features, and labels, respectively.
Given a meta-dataset of episodes $\mathcal{M}=\{\mathcal{D}^{(k)}\}_{k=1}^K$, 
the meta-learning objective is to learn the conditional predictive distribution~
$
p(y_T \mid \obin_T, \mathcal{D}_C)
.$
Intuitively, at test time, given a context dataset $\mathcal{D}_C$, and target inputs $\obin_T$, we should be able to predict the corresponding target labels $y_T$. 
For brevity, we omit superscript ${}^{(k)}$ from all per-episode quantities (e.g. $\mathcal{D}_T$, $\obin_C$, etc.)  where it is 
unambiguous and write it explicitly when distinguishing across episodes matters 
(e.g., in the meta-objective).

Neural Processes (NPs) learn  conditional $p(y_T \mid \obin_T, \mathcal{D}_C)$ across episodes in $\mathcal{M}$ while 
providing uncertainty quantification,  assuming that inputs $\obin_{[n]}$ are 
observed accurately.
In particular, NPs introduce a \emph{global} latent variable $z$ per episode to 
represent uncertainty over functions. The standard NP generative model 
is: 
\begin{align}
z \sim \mathcal{N}(0,I)~\text{(prior)}, \qquad
y_i \mid \obin_i, z \sim \mathcal{N}\!\big(f_\theta(\obin_i,z),\,\sigma_y^2\big),
\quad i \in [n],~\text{(label likelihood)}
\label{eq:np-gen}
\end{align}
where $f_\theta$ is a learnable decoder parameterized by $\theta$ and 
$\sigma_y^2$ is the observation noise variance. We write $f^\star(\cdot, z)$ for the ground-truth counterpart of $f_\theta(\cdot, z)$, i.e., the realizable optimum used in our theoretical analysis (Section~\ref{sec:theory}).
NPs utilize amortized inference, i.e., approximate the posterior on $z$ via an encoder $q_\phi(z| \cdot)$, parameterized by $\phi$; from this, NPs construct two approximate 
posteriors over $z$ given the context, $q_\phi(z \mid \mathcal{D}_C)$, 
and given context plus targets, 
$q_\phi(z \mid \mathcal{D}_C \cup \mathcal{D}_T)$, used in training.
The NP training objective is the evidence lower bound (ELBO) on the 
conditional log-likelihood:
\begin{align}
\!\!\!\log p_\theta(y_T \!\mid\! \obin_T,\mathcal{D}_C)
\!\ge
&\mathbb{E}_{q_\phi(z\mid \mathcal{D}_C\cup \mathcal{D}_T)}\!
\Big[\sum_{i\in T}\log p_\theta(y_i\!\mid\! \obin_i,z)\!\Big] 
\!-\!\mathrm{KL}\!\Big(\!q_\phi(z\!\mid\! \mathcal{D}_C\!\cup\! \mathcal{D}_T)\ \!\|\ \! q_\phi(z\!\mid\! \mathcal{D}_C)\!\Big).
\label{eq:np-elbo}
\end{align}
We denote the right-hand side of~\eqref{eq:np-elbo} as 
$\mathcal{L}_{\mathrm{ELBO}}^{(k)}(\Theta)$ when evaluated on episode 
$\mathcal{D}^{(k)}$ with split $(C^{(k)}, T^{(k)})$, where 
$\Theta = (\theta, \phi)$ collects all learnable parameters. The 
parameters are learned by maximizing the average per-episode ELBO over 
the meta-dataset:
\begin{align}
\Theta^\star \;=\; \mathop{\arg\max}_{\Theta}\; \textstyle
\frac{1}{K}\sum_{k=1}^{K} 
\mathcal{L}_{\mathrm{ELBO}}^{(k)}(\Theta).
\label{eq:np-meta-objective}
\end{align}
This objective is optimized via stochastic gradient ascent. 
At each step, a mini-batch of episodes is sampled from $\mathcal{M}$, a 
context/target split is sampled for each, and the per-episode ELBO is 
estimated with a single Monte Carlo sample from $q_\phi$ using the 
reparameterization trick. At test time, given a new episode with 
context $\mathcal{D}_C$ and target inputs $\obin_T$, predictions can be  
obtained by: (i) drawing $z \sim q_\phi(\cdot \mid \mathcal{D}_C)$, then 
(ii) for each $i \in T$, predicting $\hat{y}_i = f_\theta(\obin_i, z)$.

\section{Denoising Neural Processes (\name{})}
\label{sec:method}
Neural Processes capture two sources of uncertainty: function-level 
uncertainty through the global latent variable $z$, and output noise 
through the variance $\sigma_y^2$. However, they assume that inputs are 
observed precisely, and cannot directly model input noise. In many 
practical scenarios, such as sensor networks with known measurement 
error, the observed input $\obin_i \in \mathcal{X}$ is a noisy version 
of an underlying clean input $\nfin_i \in \mathcal{X}$. A standard NP 
applied directly to $\obin_i$ implicitly assumes the inputs are 
noise-free; the resulting model is misspecified whenever measurement 
noise is non-negligible. 

\begin{figure}
\centering
    
\includegraphics[width=0.8\textwidth]{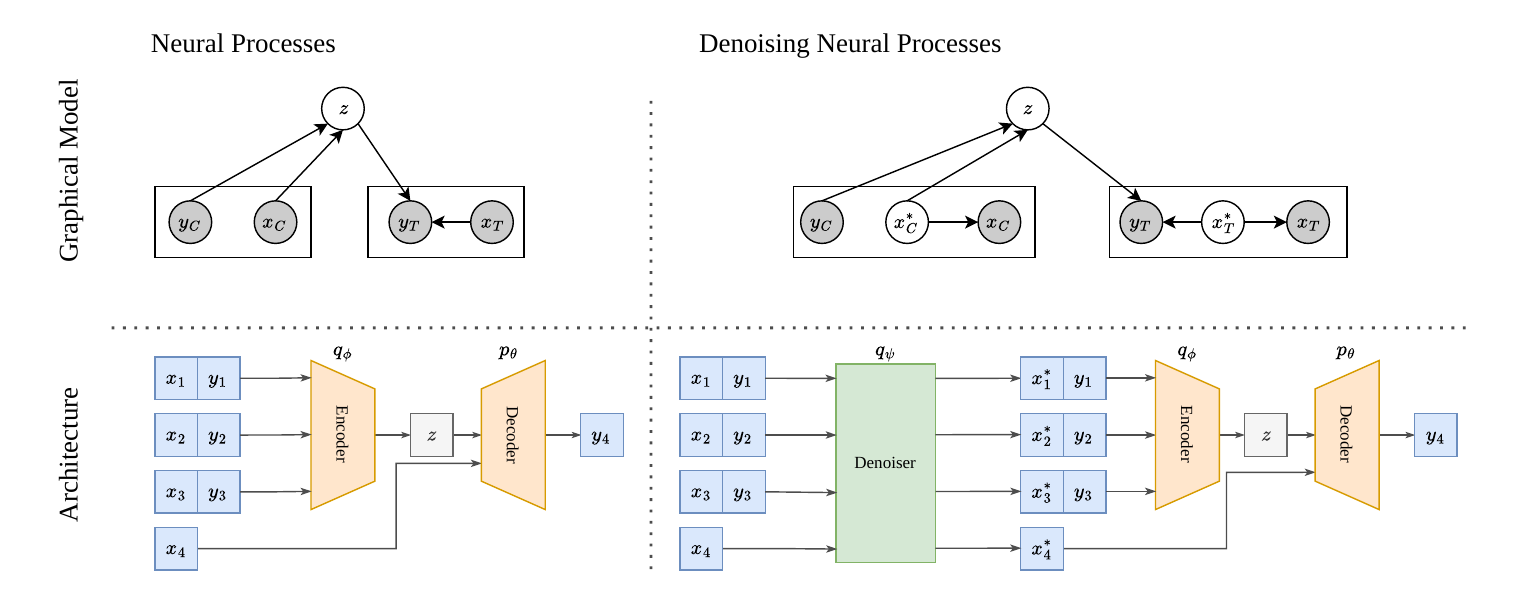}
\caption{Comparison of Neural Processes (NP) and \name{} in terms of graphical 
model (top) and architecture (bottom). In the graphical model, shaded 
nodes represent observed variables. Standard NPs assume that inputs 
are observed noise-free, an assumption violated in many real-world 
settings where measurements are inherently corrupted by noise. 
\name{} addresses this by treating noise-free inputs as latent variables 
$\nfin$ that must be inferred from noisy observations $\obin$, while 
preserving the standard NP structure of context and target sets with 
a global latent $z$ that captures function-level uncertainty. In the 
architecture, \name{} extends the standard NP encoder-decoder pipeline 
by prepending a denoising component (shown in green) that recovers 
noise-free input estimates from noisy inputs before passing them to the 
encoder.}
\label{fig:graphical_model}
\end{figure}

We address this by establishing a 
data-generative process that explicitly accounts for noisy inputs and 
propose a methodology for handling them within the Neural Process 
framework.
We adopt the same episodic setup as in Section~\ref{sec:background}: a 
meta-dataset $\mathcal{M} = \{\mathcal{D}^{(k)}\}_{k=1}^K$ of episodes, 
each split into context and target sets $C, T$. The key difference is 
that we assume that episodes $\mathcal{D} = \{(\obin_i, y_i)\}_{i=1}^n$  comprise \emph{noisy} input--output pairs: each $\obin_i$ is a 
corrupted observation of an unobserved noise-free input $\nfin_i$. 
We extend the NP generative process~\eqref{eq:np-gen} by treating noise-free 
inputs as latent variables as follows: 
\begin{equation}
\nfin_i \sim p(\nfin_i),\qquad 
\obin_i \mid \nfin_i \sim \mathcal{N}(\nfin_i,\, \Gamma), \qquad 
z \sim \mathcal{N}(0,I), \qquad
y_i \mid \nfin_i, z \sim \mathcal{N}\!\big(f_\theta(\nfin_i, z),\, \sigma_y^2\big),
\label{eq:dnp-gen}
\end{equation}
where $\Gamma \succ 0$ is the known measurement-noise covariance. The joint model is
\begin{equation}
p(\mathcal{D}, \nfin_{[n]}, z)
= \textstyle p(z) \prod_{i=1}^n p(\nfin_i)\, p(\obin_i \mid \nfin_i)\,
p_\theta(y_i \mid \nfin_i, z).
\label{eq:dnp-joint}
\end{equation}
The graphical model contains a \emph{global} latent $z$ (function 
uncertainty) and \emph{local} latents $\nfin_i$ (measurement 
uncertainty); see Figure~\ref{fig:graphical_model}. The prior 
$p(\nfin_i)$ can be chosen to reflect domain knowledge, e.g., uniform 
on a bounded region or a broad Gaussian.

The posterior $p(\nfin_{[n]}, z \mid \mathcal{D})$ is intractable 
in general. We introduce a structured variational family that mirrors 
the conditional dependence in the generative model: a denoiser $q_\psi$, 
parameterized by $\psi$, that infers noise-free inputs from noisy ones, and 
an NP-style latent encoder $q_\phi$ that infers the function-level 
latent given noise-free inputs and observations:
\begin{align}
q_{\psi,\phi}(\nfin_{[n]}, z \!\mid\! \mathcal{D})
 = q_\psi(\nfin_{[n]} \!\mid\!\obin_{[n]})\;q_\phi(z \mid \nfin_{[n]}, y_{[n]}),
\quad
q_\psi(\nfin_{[n]} \!\mid\! \obin_{[n]})
= \textstyle\prod_{i=1}^n q_\psi(\nfin_i \!\mid\! \obin_i).
\label{eq:var-family}
\end{align}
The factorization $q_\psi(\nfin_i \mid \obin_i)$ reflects an architectural 
choice: each noise-free input is inferred from its corresponding noisy 
measurement (and optionally a known noise level $\Gamma$), 
independently across observations. The latent encoder $q_\phi$ shares 
the form of the standard NP variational posterior~\eqref{eq:np-elbo} 
but now takes \emph{denoised} inputs $\nfin_{[n]}$ rather than the noisy 
$\obin_{[n]}$.

Combining the variational family~\eqref{eq:var-family} with the joint model~\eqref{eq:dnp-joint} and applying Jensen's inequality yields the evidence lower bound (ELBO) on the conditional log-likelihood of the labels given the noisy inputs (full derivation in Appendix~\ref{app:elbo}):
\begin{align}
&\log p(y_{[n]} \mid \obin_{[n]})
\ \ge\  \mathbb{E}_{q_{\psi,\phi}}
\Big[
\underbrace{\textstyle \sum_{i=1}^n \log p_\theta(y_i \mid \nfin_i, z)}_{\substack{\text{predictive}\\\text{reconstruction}}}
+
\underbrace{\textstyle\sum_{i=1}^n \log p(\obin_i \mid \nfin_i)}_{\substack{\text{measurement}\\\text{consistency}}}
\Big] \notag\\
&\quad-
\underbrace{\mathrm{KL}\!\Big(q_\psi(\nfin_{[n]} \mid \obin_{[n]}) \;\|\; p(\nfin_{[n]})\Big)}_{\substack{\text{denoiser}\\\text{regularization}}}
-
\underbrace{\mathbb{E}_{q_\psi}\mathrm{KL}\!\Big(q_\phi(z \mid \nfin_{[n]}, y_{[n]}) \;\|\; p(z)\Big)}_{\substack{\text{latent}\\\text{regularization}}}.
\label{eq:dnp-elbo-full}
\end{align}
This bound has three interpretable components: (i) a predictive 
reconstruction term that scores $y$ under \emph{denoised} inputs, 
(ii) a measurement-consistency term $\log p(\obin_i \mid \nfin_i)$ that 
ties the denoiser to the noise model, and (iii) two KL regularizers 
that prevent the denoiser and the latent encoder from drifting too far 
from their priors.

\name{} extends the NP parameter set with a denoiser network: 
$\Theta = (\theta, \phi, \psi)$, where $\theta$ parameterizes the 
decoder, $\phi$ the latent encoder, and $\psi$ the denoiser variational 
posterior $q_\psi$. For meta-learning, the quantity of interest is the conditional likelihood of the targets given the context, $\log p(y_T \mid \obin_{[n]}, y_C)$. Applying the same bound to this conditional replaces the prior $p(z)$ in~\eqref{eq:dnp-elbo-full} with the conditional prior $p(z \mid \nfin_C, y_C)$, and restricts the predictive reconstruction term to the target set. Since this conditional prior is intractable, we follow~\citep{np} and approximate it with the variational posterior $q_\phi(z \mid \nfin_C, y_C)$, which yields the training objective:
\begin{align}
\label{eq:dnp-elbo-ct}
\begin{split}
\!\!\!\!\!\!\!\!&\mathcal{L}_{\mathrm{ELBO}}^{(k)}(\Theta)
=
\mathbb{E}_{q_\psi(\nfin_{[n]} \mid \obin_{[n]})}
  \Bigg[\sum_{i=1}^n \log p(\obin_i \mid \nfin_i)\Bigg]
- \mathrm{KL}\!\Big(
    q_\psi(\nfin_{[n]}\! \mid\! \obin_{[n]})
    \;\|\;
    p(\nfin_{[n]})
  \Big)\\
&
+\mathbb{E}_{q_\psi(\nfin_{[n]} \mid \obin_{[n]})}\!
\Bigg[\!
\mathbb{E}_{q_\phi(z \mid \nfin_{[n]}, y_{[n]})}
\!\sum_{i\in T} \!\log p_\theta(y_i \mid \nfin_i, z)\!-\! \mathrm{KL}\!\Big(
    q_\phi(z \!\mid\! \nfin_{[n]}, y_{[n]})
    \;\|\;
    q_\phi(z \!\mid\! \nfin_C, y_C)
  \!\Big)
\!\Bigg].
\end{split}
\end{align}
 We learn $\Theta$ by maximizing the average per-episode ELBO over the 
meta-dataset, mirroring the NP meta-objective~\eqref{eq:np-meta-objective}. 
We optimize this objective by stochastic gradient ascent with Adam: at 
each step, we sample a mini-batch of episodes from $\mathcal{M}$, sample 
a context/target split for each, and estimate the per-episode ELBO with 
a single Monte Carlo sample from each variational distribution 
($q_\psi$ and $q_\phi$) using the reparameterization trick. Gradients 
flow through both sampling steps to update $\Theta$.
At test time, given a new episode with noisy context 
$(\obin_C, y_C)$ and noisy target inputs $\obin_T$, we predict 
$y_T$ by ancestral sampling: 
(i) draw $\nfin_{[n]} \sim q_\psi(\cdot \mid \obin_{[n]})$, 
(ii) draw $z \sim q_\phi(\cdot \mid \nfin_C, y_C)$, and 
(iii) for each $i \in T$, predict $\hat{y}_i = f_\theta(\nfin_i, z)$. 
Note that the denoising step (i) departs from standard NP 
inference. If some inputs, or individual input dimensions, are known to be noise-free, the denoiser is skipped for them and the clean values are passed directly to the encoder and decoder.

\section{Theoretical Guarantees}
\label{sec:theory}
In this section, we analyze \name{} theoretically and show that it provably outperforms noise-unaware NP baselines in the presence of input noise. We proceed in four steps. First, under a Gaussian prior, the posterior for the unknown noise-free input \(\nfin\) is Gaussian (Proposition~\ref{prop:gaussian-posterior}).  Second, we derive the Bayes-optimal predictive distribution $p^\star(y\mid\obin, z)$ under noisy inputs and characterize how input uncertainty inflates predictive variance (Equation \ref{eq:bayes-var}).  Third, we compare two predictive families: the \name{} family, which marginalizes over a learned posterior on $\nfin$, and the plug-in family used by NP, which treats $\obin$ as if it were noise-free. We prove that any plug-in model with fixed observation variance $\sigma_y^2$ incurs an irreducible KL gap to $p^\star$ whenever input uncertainty propagates through the ground-truth function (Theorem~\ref{thm:plugin-kl}), while \name{} can close this gap entirely under realizability (Theorem~\ref{thm:realizability}).  Fourth, we quantify how the plug-in gap scales with posterior input uncertainty: a moment expansion (Lemma~\ref{lem:moment-expansion}) yields explicit KL lower bounds in the small-noise regime (Theorem~\ref{thm:scaling-kl}), which simplifies to a clean $\tau^4 \|\nabla f^\star\|^4 / (4\sigma_y^4)$ scaling under isotropic posterior covariance (Corollary~\ref{cor:scaling-isotropic}). All proofs are deferred to Appendix~\ref{app:theory-proofs}. We use a Gaussian prior on $\nfin$ as a concrete example; the proof and 
an extension to a uniform prior are in Appendix~\ref{app:props}.

\begin{proposition}[Gaussian posterior under Gaussian corruption]
\label{prop:gaussian-posterior}
Under~\eqref{eq:dnp-gen} with a Gaussian prior $\nfin \sim \mathcal{N}(\mu, \Sigma)$, 
$\Sigma \succ 0$, the posterior over the noise-free input is Gaussian, $p(\nfin\mid \obin) = \mathcal{N}\!\big(\nfin \mid 
\mu_{\mathrm{post}}(\obin),\, \Sigma_{\mathrm{post}}\big),
$
with
$\mu_{\mathrm{post}}(\obin)
= (\Sigma^{-1}+\Gamma^{-1})^{-1}(\Sigma^{-1}\mu+\Gamma^{-1}\obin),
\notag
\Sigma_{\mathrm{post}}
= (\Sigma^{-1}+\Gamma^{-1})^{-1}.$
\end{proposition}

A bounded-domain version, obtained by replacing the Gaussian prior 
with a uniform prior on a hyper-rectangle, yields a truncated 
multivariate normal posterior; see Appendix~\ref{app:props}.

{\bf Bayes-Optimal Predictive Conditional}. Given Proposition~\ref{prop:gaussian-posterior}, the Bayes-optimal 
predictive conditional under noisy inputs is $p^\star(y\mid \obin,z)
=
\int \mathcal{N}\!\big(y\mid f^\star(\nfin,z),\sigma_y^2\big)
\,p(\nfin\mid \obin)\,d\nfin.$ If $X\sim p(\nfin\mid \obin)$ and 
$\varepsilon\sim \mathcal{N}(0,\sigma_y^2)$ is independent of $X$,
then $Y=f^\star(X,z)+\varepsilon$ has conditional mean and variance
\begin{align}
\mu_\star(\obin,z)
&:= \mathbb{E}[Y\mid \obin,z]
= \mathbb{E}\!\left[f^\star(X,z)\mid \obin\right],
\\
v_\star(\obin,z)
&:= \mathrm{Var}(Y\mid \obin,z)
= \sigma_y^2+\mathrm{Var}\!\left(f^\star(X,z)\mid \obin\right).
\label{eq:bayes-var}
\end{align}
Equation~\eqref{eq:bayes-var} shows explicitly how input uncertainty 
inflates predictive variance. Since $p^\star(y\mid\obin,z)$ 
requires integrating over the intractable posterior $p(\nfin \mid\obin,z)$,
we instead utilize the variational posterior $q_\psi(\nfin\mid \obin)$.
The difference between \name{} and noise-unaware NP is most transparent at the level of
conditional predictive distributions given the episode representation.
Accordingly, throughout this section we condition on the episode latent $z$
(random for NP).
We assume the scalar-output generative model as in Section~\ref{sec:method}.

The \name{} predictive family (conditional on $z$) is
\begin{equation}
q_{\theta,\psi}^{\mathrm{\name{}}}(y\mid \obin,z)
=
\int \mathcal N\!\big(y\mid f_\theta(\nfin,z),\sigma_y^2\big)\,q_\psi(\nfin\mid \obin)\,d\nfin.
\label{eq:\name{}-class}
\end{equation}
In contrast, a noise-unaware NP baseline that plugs in $\obin$ as if it were noise-free has conditional form
\begin{equation}
q_{\theta}^{\mathrm{plug}}(y\mid \obin,z)
=
\mathcal N\!\big(y\mid m_\theta(\obin,z),\sigma_y^2\big),
\label{eq:plugin-class}
\end{equation}
where $m_\theta$ is a measurable mean function. For NPs, \eqref{eq:plugin-class} is the conditional decoder after fixing the global latent $z$.

\begin{theorem}[Irreducible error in NPs]
\label{thm:plugin-kl}
Fix $(\obin,z)$ and let \(
P_{\obin,z}:=p^\star(\cdot\mid \obin,z).\) For any measurable mean function $m(\obin,z)$,
\begin{align}
\mathrm{KL}\!\big(P_{\obin,z}\ \|\ \mathcal N\!\big(m(\obin,z),\sigma_y^2\big)\big)
\ge 
\frac{\big(\mu_\star(\obin,z)-m(\obin,z)\big)^2}{2\sigma_y^2}\notag +
\frac12\left(
\frac{v_\star(\obin,z)}{\sigma_y^2}
-1
-\log\frac{v_\star(\obin,z)}{\sigma_y^2}
\right).
\label{eq:kl-lb-general}
\end{align}
In particular,
\begin{equation}
\inf_{m}
\mathrm{KL}\!\big(
P_{\obin,z}\ \|\ \mathcal N(m,\sigma_y^2)
\big)
\ge
\frac12\left(
\frac{v_\star(\obin,z)}{\sigma_y^2}
-1
-\log\frac{v_\star(\obin,z)}{\sigma_y^2}
\right),
\label{eq:kl-lb-varonly}
\end{equation}
and the right-hand side is strictly positive whenever
$\mathrm{Var}\!\big(f^\star(X,z)\mid \obin\big) > 0.$
\end{theorem}

Equation~\eqref{eq:kl-lb-varonly} shows that NP-based models incur an irreducible KL gap whenever input uncertainty propagates through $f^\star$. We now show that \name{}, in contrast, can close this gap entirely because of the noise awareness.

\begin{theorem}[Noise-aware realizability of DnNP]
\label{thm:realizability}
Assume there exist parameters $(\theta^\star,\psi^\star)$ such that
\begin{align}
f_{\theta^\star}(\nfin,z)= &f^\star(\nfin,z)
\quad\text{for all }(\nfin,z),\notag\\
q_{\psi^\star}(\nfin\mid \obin)= &p(\nfin\mid \obin)
\quad\text{for all }\obin.
\label{eq:realizability-assumptions}
\end{align}
Then the noise-aware predictor coincides with the Bayes predictor:
\begin{equation}
q_{\theta^\star,\psi^\star}^{\mathrm{\name{}}}(y\mid \obin,z)
=
p^\star(y\mid \obin,z)
\quad\text{for all }(\obin,z).
\label{eq:realizability-pointwise}
\end{equation}
Consequently,
\begin{equation}
\inf_{\theta,\psi}
\mathbb E_{\obin,Z}
\Big[
\mathrm{KL}\!\big(
p^\star(\cdot\mid \obin,Z)
\ \|\ 
q_{\theta,\psi}^{\mathrm{\name{}}}(\cdot\mid \obin,Z)
\big)
\Big]
=
0.
\label{eq:realizability-kl}
\end{equation}
\end{theorem}

Theorem~\ref{thm:realizability} shows that the Bayes predictor under noisy inputs belongs to the \name{} class. We now quantify how the plug-in KL gap from Theorem~\ref{thm:plugin-kl} scales with posterior input uncertainty in the small-noise regime. We first establish a moment expansion for the Bayes predictive moments, then combine it with Theorem~\ref{thm:plugin-kl} to obtain explicit KL lower bounds.

\begin{lemma}[Small-noise moment expansion]
\label{lem:moment-expansion}
Assume $f^\star(\cdot,z)$ is twice continuously differentiable with globally Lipschitz Hessian. As $\|\Sigma_{\mathrm{post}}\|\to 0$,
\begin{align}
\mu_\star(\obin,z)
&=
f^\star(\mu_{\mathrm{post}}(\obin),z)
+
\frac12 \mathrm{tr}\!\big(\nabla_{\nfin}^2 f^\star(\mu_{\mathrm{post}}(\obin),z)\,\Sigma_{\mathrm{post}}\big)
+
O(\|\Sigma_{\mathrm{post}}\|^{3/2}),
\label{eq:scaling-mean}
\\
v_\star(\obin,z)-\sigma_y^2
&=
(\nabla_{\nfin} f^\star(\mu_{\mathrm{post}}(\obin),z))^\top \Sigma_{\mathrm{post}}\,\nabla_{\nfin} f^\star(\mu_{\mathrm{post}}(\obin),z)
+
O(\|\Sigma_{\mathrm{post}}\|^2).
\label{eq:scaling-var}
\end{align}
\end{lemma}
This lemma shows the posterior mean and variance of the ground truth function \(f^\star\).

\begin{theorem}[NP KL gap in the small-noise regime]
\label{thm:scaling-kl}
Under the assumptions of Lemma~\ref{lem:moment-expansion}, as $\|\Sigma_{\mathrm{post}}\|\to 0$,
\begin{align}
\begin{split}
&\mathrm{KL}\big(P_{\obin,z}\ \|\ \mathcal N\!\big(f^\star(\mu_{\mathrm{post}}(\obin),z),\,\sigma_y^2\big)\big)
\ge\;
\frac{1}{8\sigma_y^2}
\big(\mathrm{tr}(\nabla_{\nfin}^2 f^\star(\mu_{\mathrm{post}}(\obin),z)\,\Sigma_{\mathrm{post}})\big)^2 \\
&+
\frac{1}{4\sigma_y^4}
\big(\nabla_{\nfin} f^\star(\mu_{\mathrm{post}}(\obin),z)^\top \Sigma_{\mathrm{post}}\,\nabla_{\nfin} f^\star(\mu_{\mathrm{post}}(\obin),z)\big)^2
+
o(\|\Sigma_{\mathrm{post}}\|^2).
\end{split}
\label{eq:scaling-kl}
\end{align}
Moreover, after optimizing over the predictive mean,
\begin{equation}
\inf_{m}
\mathrm{KL}\!\big(
P_{\obin,z}\ \|\ \mathcal N(m,\sigma_y^2)
\big)
\ge
\frac{1}{4\sigma_y^4}
\big(\nabla_{\nfin} f^\star(\mu_{\mathrm{post}}(\obin),z)^\top \Sigma_{\mathrm{post}}\,\nabla_{\nfin} f^\star(\mu_{\mathrm{post}}(\obin),z)\big)^2
+
o(\|\Sigma_{\mathrm{post}}\|^2).
\label{eq:scaling-kl-irreducible}
\end{equation}
\end{theorem}
Then, in small noise regime, we have the concrete KL lower bound for NP and the optimal Bayes predictor. Moreover, if we consider an isotropic posterior covariance, we have the following corollary.

\begin{corollary}[Isotropic posterior covariance]
\label{cor:scaling-isotropic}
Under the assumptions of Theorem~\ref{thm:scaling-kl}, if $\Sigma_{\mathrm{post}}=\tau^2 I_d$, then the irreducible term in~\eqref{eq:scaling-kl-irreducible} scales as
$
\frac{\tau^4}{4\sigma_y^4}
\big\|\nabla_{\nfin} f^\star(\mu_{\mathrm{post}}(\obin),z)\big\|_2^4
+
o(\tau^4).
$
\end{corollary}
Thus, in the small noise regime, the irreducible error is related to the gradient of \(f^\star\) at the ground truth \(x^*\), the input noise variance, and the output noise variance.

\section{Experiments}
\label{sec:experiment}
To evaluate the effectiveness of \name{} in handling  noisy inputs, we conduct experiments comparing it against multiple Neural Processes variants on four datasets. We begin by describing the datasets and baseline methods, then introduce the experimental configuration, and conclude with a discussion of the results.

{\bf Data}. Table~\ref{tbl:dataset} in the appendix summarizes the datasets used in our experiments. The synthetic dataset comprises 1D regression tasks with functions sampled from a GP with an RBF kernel. We also include MNIST ~\citep{mnist} image completion, a widely-used benchmark in Neural Processes evaluations. Additionally, we apply our method to two real-world spatial interpolation datasets: an Air Quality dataset of PM$_{2.5}$ measurements that we collected from a network of urban sensors in Brookline, MA, which we make publicly available at \url{https://hdl.handle.net/2047/D20881761}, and a publicly available dataset of daily sea-level pressure observations from NOAA weather stations across California~\cite{noaa_ncei}, obtained via the Meteostat library~\cite{meteostat}. Both datasets contain real-world sensor measurements with inherent noise, demonstrating the importance of accounting for input uncertainty in practical applications.

{\bf Methods}. We compared \name{} with standard Neural Processes (NP) ~\citep{np}, Conditional Neural Processes (CNP)~\citep{cnp}, Attentive Neural Processes (ANP)~\citep{anp}, and Convolutional Conditional Neural Processes (ConvCNP)~\citep{convcnp}. To assess the impact of input noise, we train an additional baseline on noise-free data for synthetically corrupted datasets (1D GP and MNIST), representing the best achievable performance without input noise. To simulate input uncertainty, we add zero-mean Gaussian noise to the model inputs with varying noise levels. Noise is applied independently across input dimensions and data points, modeling realistic measurement scenarios in which each sensor reading is affected by independent noise.

 {\bf Metrics}.
We evaluate model performance on the target set using the posterior predictive, with latent variables marginalized by Monte Carlo: we draw $K$ samples from the variational posteriors, obtaining predictive means and standard deviations $(\mu_k, \sigma_k)$, and compute $\mathrm{NLL} = -\frac{1}{|T|}\sum_{i \in T} \log \frac{1}{K}\sum_{k=1}^{K} \mathcal{N}(y_i \mid \mu_{k,i}, \sigma_{k,i}^2)$ and $\mathrm{RMSE} = \sqrt{\frac{1}{|T|}\sum_{i\in T} (y_i - \bar{\mu}_i)^2}$, where $\bar{\mu}_i = \frac{1}{K}\sum_k \mu_{k,i}$.
 We compare our method against two neural process (NP) baselines. The first baseline is an NP trained on noise-free data (prior to noise injection). The second baseline is an NP trained directly on noisy inputs, representing the setting in which input uncertainty is ignored during modeling.
  We adopt the same backbone architecture for \name{} as for NP, with the addition of a denoising component. The model is trained using the loss defined in Equation~\ref{eq:dnp-elbo-ct}. Hyper-parameters, including the weighting coefficient $\beta$ that controls the strength of the denoiser KL regularizer relative to the predictive reconstruction term, are selected via grid search. The model achieving the lowest validation NLL is selected and evaluated on the test set, where it is compared against all baselines.
  
\subsection{1D Synthetic Functions}

\begin{figure}[t]
\centering
\includegraphics[width=0.8\columnwidth]{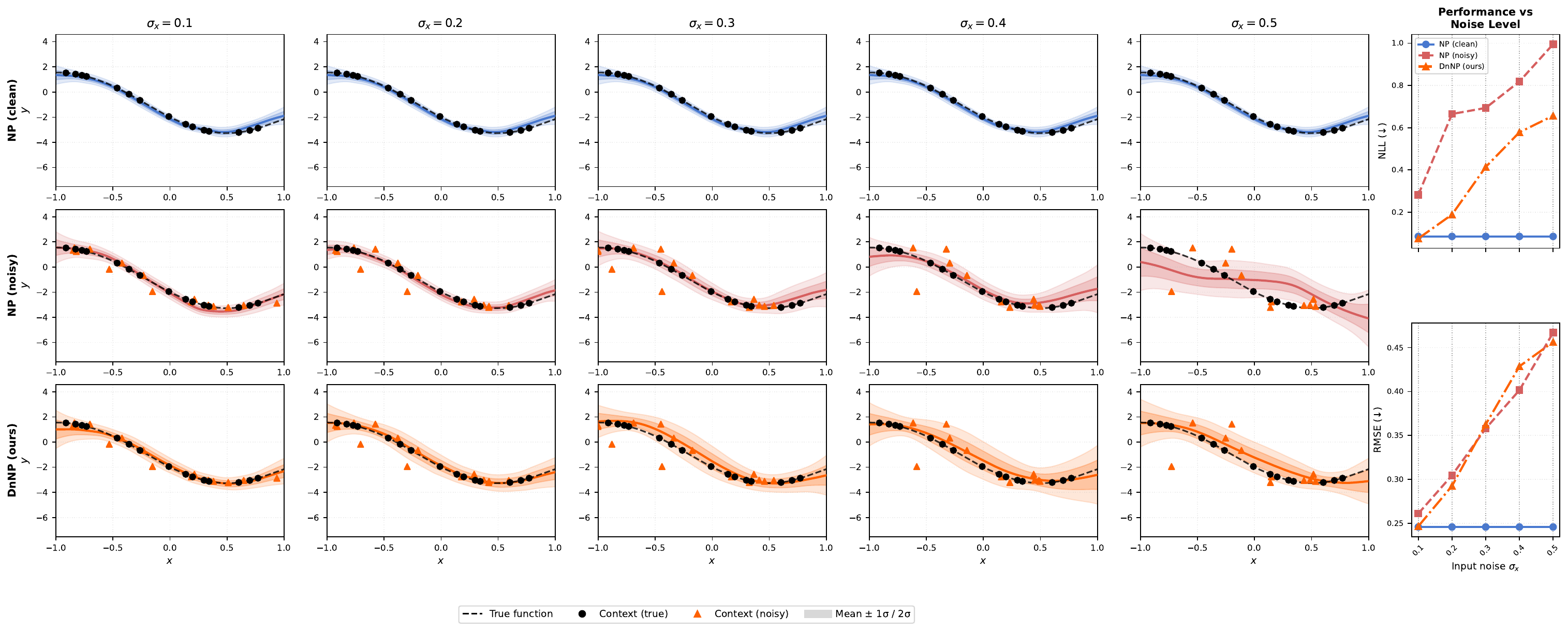}
\caption{Comparison of Clean NP, Noisy NP, and \name{} on 1D GP regression across noise levels  $\sigma_x \in \{0.1,0.2,0.3,0.4,0.5\}$. Shaded regions show the predictive mean $\pm \sigma$ and $\pm 2\sigma$ uncertainty bands. Black circles indicate true context inputs and orange triangles show the corrupted observed input. As input noise increases, Noisy NP produces increasingly biased mean predictions, whereas \name{} maintains closer agreement with the true function. The NLL sweep (right) confirms this quantitatively, showing that \name{} achieves consistently lower NLL than Noisy NP across all noise levels.}
\label{fig:toy}
\end{figure}

We evaluated on synthetic 1D regression tasks where functions are sampled from a Gaussian Process with an RBF kernel (length-scale 0.5). During both training and evaluation, context points are corrupted by additive Gaussian noise $x_C = \nfin_C + \varepsilon , \, \varepsilon \sim \mathcal{N}(0, \sigma_x ^2)$, across five noise levels $\sigma_x \in \{0.1,0.2,0.3,0.4,0.5\}$, with a separate model trained for each noise level. All models share the same encoder-decoder architecture (hidden dim 128, latent dim 64) and are trained on 10,000 GP-sampled tasks with random context size drawn uniformly from [5, 30]. Clean NP is trained and evaluated on noise-free inputs $\nfin$ and serves as a lower bound on both NLL and RMSE, while Noisy NP is trained and evaluated on corrupted inputs $x$. Figure ~\ref{fig:toy} shows that as noise increases, Noisy NP progressively drifts from the true function. The NLL sweep confirms this quantitatively, showing that \name{} achieves consistently lower NLL across all noise levels.

\name{} is also robust to misspecification of the input noise level: on the 1D benchmark with true $\sigma_x = 0.3$, training with an assumed $\sigma$ ranging from $0.5$ to $2$ times the true value changes RMSE only from $0.363$ to $0.406$ (vs.\ $0.373$ at the true $\sigma$), with NLL nearly flat ($\approx 0.47$--$0.54$), within seed variation. This is consistent with our real-world experiments, where $\sigma$ is set from manufacturer specifications rather than known exactly. When no estimate of $\sigma$ is available, $\sigma$ can be learned jointly with the model; the learned value tracks the true noise level monotonically, with a slight upward bias, and gives predictive performance close to, though not exceeding, the fixed-$\sigma$ model, where the known $\sigma$ is provided. We also experimented with more complex function classes such as periodic kernels, where all models, including Clean NP, underfit; this is consistent with prior observations that latent NPs with a global representation require more expressive architectures, such as attentive or convolutional NPs, to capture such structure~\citep{dubois2020npf}.

\subsection{MNIST Image Completion}
  We use the MNIST dataset of $28\times 28$ grayscale handwritten digits with the standard 60K/10K train/test split. We further split the test set into 5K validation and 5K test subsets (fixed random seed for reproducibility). The task is image completion: given sparse pixel observations, predict the full 784-pixel image. During training, we randomly sample 3-200 context points and use all 784 pixels as targets. At test time, we use 100 context points. Inputs are 2D pixel coordinates $(x_1, x_2) $ normalized to $[-1, 1]^2$. To simulate input uncertainty, we corrupt context inputs by adding independent Gaussian noise $\mathcal{N}(\mathbf{0}, \sigma_x^2 \mathbf{I}_2)$ where 
 $\sigma_x \in \{0.001, 0.005, 0.01, 0.05, 0.1, 0.2, 0.5\}$ in the normalized input space. These noise levels correspond to average pixel displacements of 0.02–7.4 pixels. We train separate NP and \name{} models for each noise level.

\begin{figure}[t]
\centering
\includegraphics[width=0.75\textwidth]{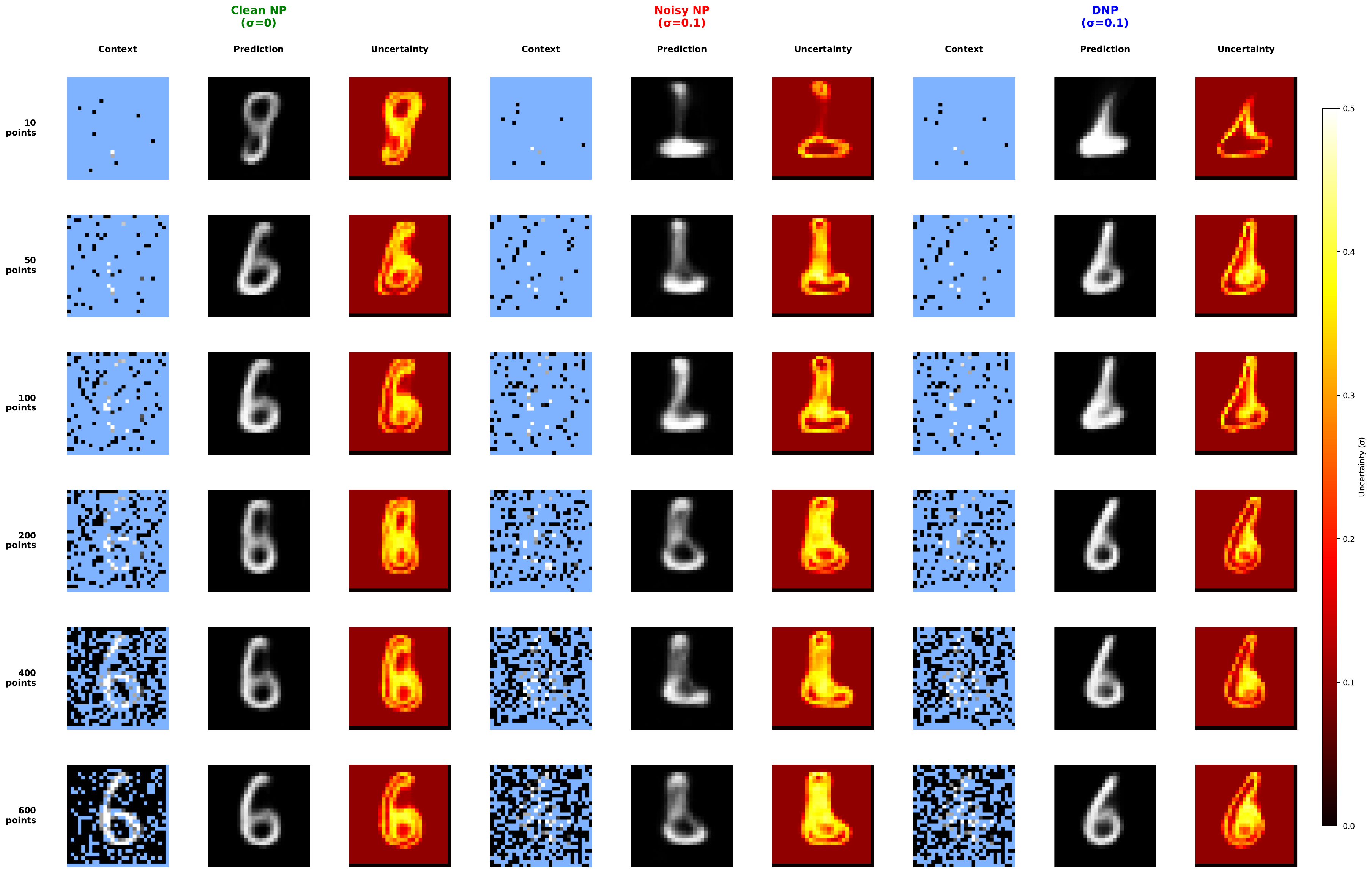}
\caption{MNIST image completion with noisy pixel coordinates ($\sigma_x=0.1$) at varying context sizes. Each row shows context, prediction, and per-pixel uncertainty for Clean NP ($\sigma_x=0$, reference), Noisy NP, and \name{}. Noisy NP fails to recover the correct digit at all context sizes, while \name{} recovers digits comparable to Clean NP, with uncertainty concentrated along edges.}
\label{fig:mnist-qualitative}
\end{figure}

Table~\ref{tab:mnist-clean} and Figure~\ref{fig:mnist-qualitative} summarize the results on MNIST image completion under input noise. \name{} outperforms Noisy NP in NLL and RMSE at every noise level, and in accuracy for $\sigma_x \leq 0.1$, demonstrating that explicitly modeling input noise leads to more accurate and better-calibrated predictions. Notably, at low to moderate noise levels, \name{} recovers performance close to the noise-free baseline. As noise increases further, accuracy degrades for all methods as expected, since highly corrupted coordinates make the spatial structure of the image increasingly difficult to recover; at the largest noise levels ($\sigma_x \geq 0.2$), accuracy drops close to chance for both models, and the two are comparable. Note that RMSE and NLL remain nearly constant across noise levels because they are dominated by pixel-intensity reconstruction, which changes little as coordinate noise grows, whereas classification accuracy depends directly on whether the spatial structure of the digit is preserved; accuracy is therefore the more discriminative metric for this task. Qualitatively, Figure~\ref{fig:mnist-qualitative} shows that Noisy NP produces a persistent shape distortion regardless of context size, whereas \name{} recovers sharper and more coherent digit structures.
\begin{table}[t]
\caption{MNIST results. Left: clean-input baselines ($\sigma_x = 0$). Right: performance under increasing input noise $\sigma_x$ for Noisy NP vs.\ \name{}; bold marks the better of the two at each noise level.}
\centering
\begin{minipage}[t]{0.32\columnwidth}
\centering
\resizebox{\textwidth}{!}{%
\begin{tabular}{lccc}
\toprule
Method & NLL $\downarrow$ & RMSE $\downarrow$ & Acc $\uparrow$ \\
\midrule
NP       & -0.0631  & 0.1413 & 0.706 \\
CNP      & -0.00571 & 0.232  & 0.319 \\
ANP      & -0.00702 & 0.190  & 0.733 \\
ConvCNP  & -0.00454 & 0.213  & 0.576 \\
\bottomrule
\end{tabular}%
}
\label{tab:mnist-clean}
\end{minipage}%
\hfill
\begin{minipage}[t]{0.66\columnwidth}
\centering
\resizebox{\textwidth}{!}{%
\begin{tabular}{llccccccc}
\toprule
& & \multicolumn{7}{c}{$\sigma_x$} \\
\cmidrule(lr){3-9}
Metric & Method & 0.001 & 0.005 & 0.01 & 0.05 & 0.1 & 0.2 & 0.5 \\
\midrule
\multirow{2}{*}{NLL $\downarrow$}
& Noisy NP & -0.0349 & -0.0398 & -0.0403 & -0.0400 & -0.0401 & -0.0393 & -0.0380 \\
& \name{}  & \textbf{-0.0616} & \textbf{-0.0630} & \textbf{-0.0624} & \textbf{-0.0637} & \textbf{-0.0646} & \textbf{-0.0637} & \textbf{-0.0628} \\
\midrule
\multirow{2}{*}{RMSE $\downarrow$}
& Noisy NP & 0.243 & 0.236 & 0.232 & 0.233 & 0.233 & 0.235 & 0.239 \\
& \name{}  & \textbf{0.148} & \textbf{0.142} & \textbf{0.145} & \textbf{0.138} & \textbf{0.134} & \textbf{0.139} & \textbf{0.143} \\
\midrule
\multirow{2}{*}{Acc $\uparrow$}
& Noisy NP & 0.205 & 0.207 & 0.221 & 0.210 & 0.215 & \textbf{0.193} & \textbf{0.157} \\
& \name{}  & \textbf{0.660} & \textbf{0.726} & \textbf{0.710} & \textbf{0.592} & \textbf{0.370} & 0.184 & 0.123 \\
\bottomrule
\end{tabular}%
}
\label{tab:mnist-noisy}
\end{minipage}
\end{table}
 
\subsection{Spatial Interpolation}
We collect hourly PM$_{2.5}$ data from 34 QuantAQ~\citep{quantaq} air quality sensors located in Brookline, Massachusetts, USA, a town of approximately 17.6~km$^2$. The dataset is publicly available at \url{https://hdl.handle.net/2047/D20881761}. The analysis period spans May 24 to September 19, 2024, and all sensors used had more than 90\% data completeness over the study period. Missing values were filled using linear imputation. We apply a log transformation to the PM$_{2.5}$ data and min-max normalization to other predictor variables such as latitude, longitude, temperature, and humidity.
The dataset is divided into three splits along both the temporal and spatial axes. Temporally, the data are sequentially partitioned into three consecutive segments for training, validation, and testing: the first 80\% for training, the next 10\% for validation, and the remaining 10\% for testing. Spatially, sensors are randomly split into three non-overlapping groups—context, target, and held-out—at approximately 55\%, 15\%, and 30\%, respectively. For the training and validation splits, the context sensors are used to construct the feature vector and the model is trained/evaluated on the target sensors. To assess generalization to unobserved locations, the test split uses context-sensor values to construct the feature vector and evaluates predictions on the held-out sensors.

Input noise for the temperature and relative humidity 
dimensions is derived from the MODULAIR\texttrademark{} sensor 
specifications~\citep{quantaq}, which report measurement 
accuracies of $\pm 0.2^\circ$C for temperature and $\pm 1.8\%$ for 
relative humidity. Table~\ref{tab:spatial_interpolation} presents results on PM$_{2.5}$ spatial interpolation on the Air Quality dataset across three feature configurations. \name{} consistently outperforms standard NPs across all feature sets in both NLL and RMSE, and both improve over the mean baseline. These results demonstrate that explicitly accounting for noisy input is particularly beneficial in spatial interpolation settings, where sensor data uncertainty is a natural and unavoidable source of noise. We further verified that the denoiser does not collapse to the identity map: its corrections to the noisy inputs are non-zero across all feature configurations and seeds, and small relative to the assumed noise level, as expected when sensor noise is small compared to the spread of each feature.

We additionally evaluate \name{} on sea-level pressure interpolation across NOAA weather stations in California. We retrieve daily observations of sea-level pressure, temperature, and relative humidity via the Meteostat library~\citep{meteostat} for Q1 2026, retaining 29 stations with $\geq 95\%$ data completeness over the 90-day period. Missing values are linearly imputed, sea-level pressure is the prediction target, and predictor variables (latitude, longitude, temperature, humidity) are min-max normalized. We split the 90 days temporally into 80\% training, 10\% validation, and 10\% test. Within each daily episode, sensors are randomly partitioned into context and target sets following the same protocol as the air quality experiment.
Input noise standard deviations for temperature and relative humidity are derived from ASOS sensor specifications, which report typical accuracies of $\pm 0.5^\circ$C for the platinum resistance thermometer and $\pm 2\%$ for the Vaisala capacitive humidity sensor. The corresponding rows of Table~\ref{tab:spatial_interpolation} report results across three input feature configurations. \name{} achieves lower RMSE than the NP in all three settings and lower NLL in two of three, with the largest gains under (Lat, Lon, RH) and (Lat, Lon, Temp, RH), where humidity enters as a noisy input. In the (Lat, Lon, Temp) setting, the two models are statistically tied in NLL ($-1.240 \pm 0.032$ vs.\ $-1.239 \pm 0.021$), while \name{} still reduces RMSE.

\newcommand{\best}[1]{\textbf{\boldmath$#1$}}

\begin{table}[t]
\centering
\caption{Spatial interpolation results on two real-world datasets, reported as mean $\pm$ standard deviation over 5 seeds. Mean RMSE is the constant-prediction baseline (predicting the dataset mean). Bold marks the better mean of NP vs.\ \name{} per metric within each row.}
\label{tab:spatial_interpolation}
\setlength{\tabcolsep}{5pt}
\renewcommand{\arraystretch}{1.2}
\resizebox{\textwidth}{!}{%
\begin{tabular}{llc cc cc}
\toprule
\multirow{2}{*}{\textbf{Dataset}}
  & \multirow{2}{*}{\textbf{Feature Set}}
  & \textbf{Mean}
  & \multicolumn{2}{c}{\textbf{NP}}
  & \multicolumn{2}{c}{\textbf{\name{} (ours)}} \\
\cmidrule(lr){4-5} \cmidrule(lr){6-7}
  & & \textbf{RMSE $\downarrow$}
  & \textbf{NLL $\downarrow$} & \textbf{RMSE $\downarrow$}
  & \textbf{NLL $\downarrow$} & \textbf{RMSE $\downarrow$} \\
\midrule
\multirow{3}{*}{\shortstack[l]{PM2.5\\(Air Quality)}}
& Lat, Lon, RH       & 0.776 & $1.098 \pm 0.273$ & $0.457 \pm 0.036$ & \best{0.296 \pm 0.078} & \best{0.320 \pm 0.023} \\
& Lat, Lon, Temp     & 0.776 & $1.038 \pm 0.187$ & $0.439 \pm 0.030$ & \best{0.363 \pm 0.089} & \best{0.337 \pm 0.055} \\
& Lat, Lon, RH, Temp & 0.776 & $0.925 \pm 0.161$ & $0.430 \pm 0.032$ & \best{0.351 \pm 0.130} & \best{0.348 \pm 0.062} \\
\midrule
\multirow{3}{*}{\shortstack[l]{Pressure\\(NOAA CA 2026)}}
& Lat, Lon, Temp     & 0.108 & \best{-1.240 \pm 0.032} & $0.052 \pm 0.008$ & $-1.239 \pm 0.021$     & \best{0.036 \pm 0.006} \\
& Lat, Lon, RH       & 0.108 & $-1.091 \pm 0.064$      & $0.076 \pm 0.008$ & \best{-1.229 \pm 0.036} & \best{0.039 \pm 0.010} \\
& Lat, Lon, Temp, RH & 0.108 & $-1.171 \pm 0.031$      & $0.064 \pm 0.005$ & \best{-1.205 \pm 0.067} & \best{0.048 \pm 0.015} \\
\bottomrule
\end{tabular}%
}
\end{table}

\section{Conclusion}
In this paper, we introduced \name{}, a new member of the Neural Process family that explicitly handles noisy input coordinates through a variational denoising mechanism. We evaluated \name{} on 1D GP regression, MNIST image completion, and two real-world PM$_{2.5}$ and sea-level pressure spatial interpolation datasets, achieving lower or comparable NLL and lower RMSE across different settings, with up to 30\% RMSE reduction on the air quality dataset. We further provided theoretical guarantees showing that the Bayes-optimal predictor under noisy inputs belongs to the \name{} model class, while plug-in models that treat noisy inputs as clean incur an irreducible KL divergence gap. Natural future directions include combining \name{} with methods that handle noisy labels~\citep{noisy_y_np}, particularly relevant for air quality datasets where both input covariates and target measurements are corrupted, and extending the framework to noise that is correlated across observations.

\begin{ack}
This work was supported by the National Institutes of Health (NIH) under grant 5R01HL167072, the National Science Foundation (NSF) under grant 2414652, and the iSUPER Impact Engine at Northeastern University.
\end{ack}

\newpage

\bibliography{refrences}
\bibliographystyle{plain}

\newpage
\appendix
\section{Limitations} 
\label{app:limitations}

DnNP assumes measurement noise that is independent across observations, encoded by the pointwise denoiser $q_\psi(x^*_i \mid x_i)$; spatially correlated sensor bias, documented in both of our application domains, is not captured, and a set encoder approximating the joint posterior over clean inputs is a natural extension. The method also relies on a user-specified noise model and prior over clean inputs, which reduces the purely data-driven character of NPs and opens the door to misspecification, although DnNP is robust to moderate misspecification of $\sigma$. Finally, DnNP inherits the capacity of its backbone; since the denoiser is architecture-agnostic, combining it with more expressive models such as TNPs is a promising direction.

\section{Broader Impact}
\label{app:broader_impact}
 \name{} is designed for settings where input measurements are inherently noisy, a common condition in real-world sensor networks. Our spatial interpolation experiments on urban air quality data demonstrate that explicitly modeling input noise improves prediction accuracy in environmental monitoring applications. Improved interpolation in this setting can support exposure assessment, environmental justice analyses, and public health decision-making, particularly in cities that rely on sparse, low-cost sensor deployments where measurement uncertainty is unavoidable. More broadly, principled handling of input uncertainty is relevant to many scientific and engineering domains that depend on noisy measurements, including geosciences and climate modeling. We recommend that practitioners deploying \name{} validate the assumed input-noise model against sensor specifications or held-out calibration data.

\section{Posterior Over Noise-Free Inputs}
\label{app:props}

Here we provide the proofs for the posterior propositions including the Gaussian prior and a truncated uniform prior.

\subsection{Proof of Proposition~\ref{prop:gaussian-posterior}}

\begin{proof}
By Bayes' rule,
\[
p(\nfin\mid \obin)\propto p(\obin\mid \nfin)\,p(\nfin).
\]
Using the generative model from Section~\ref{sec:method},
\[
p(\obin\mid \nfin)\propto
\exp\!\Big(-\tfrac12(\obin-\nfin)^\top\Gamma^{-1}(\obin-\nfin)\Big),
\qquad
p(\nfin)\propto
\exp\!\Big(-\tfrac12(\nfin-\mu)^\top\Sigma^{-1}(\nfin-\mu)\Big).
\]
Therefore,
\[
p(\nfin\mid \obin)
\propto
\exp\!\Big(
-\tfrac12\Big[(\nfin-\mu)^\top\Sigma^{-1}(\nfin-\mu)
+
(\nfin-\obin)^\top\Gamma^{-1}(\nfin-\obin)\Big]
\Big).
\]
Expanding the quadratic form in $\nfin$ yields
\[
(\nfin-\mu)^\top\Sigma^{-1}(\nfin-\mu)
+
(\nfin-\obin)^\top\Gamma^{-1}(\nfin-\obin)
=
\nfin^\top(\Sigma^{-1}+\Gamma^{-1})\nfin
-
2(\Sigma^{-1}\mu+\Gamma^{-1}\obin)^\top \nfin
+
\mathrm{const}.
\]
Set
\[
\Lambda_{\mathrm{post}}:=\Sigma^{-1}+\Gamma^{-1},
\qquad
\Sigma_{\mathrm{post}}:=\Lambda_{\mathrm{post}}^{-1},
\qquad
h:=\Sigma^{-1}\mu+\Gamma^{-1}\obin.
\]
Then
\[
\nfin^\top\Lambda_{\mathrm{post}}\nfin-2h^\top \nfin
=
(\nfin-\mu_{\mathrm{post}}(\obin))^\top
\Lambda_{\mathrm{post}}
(\nfin-\mu_{\mathrm{post}}(\obin))
-
\mu_{\mathrm{post}}(\obin)^\top
\Lambda_{\mathrm{post}}
\mu_{\mathrm{post}}(\obin),
\]
where
\[
\mu_{\mathrm{post}}(\obin):=\Sigma_{\mathrm{post}}h.
\]
Hence
\[
p(\nfin\mid \obin)
\propto
\exp\!\Big(
-\tfrac12
(\nfin-\mu_{\mathrm{post}}(\obin))^\top
\Sigma_{\mathrm{post}}^{-1}
(\nfin-\mu_{\mathrm{post}}(\obin))
\Big),
\]
which is the Gaussian density stated in Proposition~\ref{prop:gaussian-posterior}.
\end{proof}

\subsection{Bounded-Domain Extension}

\begin{proposition}[Uniform box prior + Gaussian corruption]
\label{prop:trunc-mvn}
Let $\nfin\in\mathbb R^d$ have a uniform prior on the box
\[
\mathcal B=[a_1,b_1]\times\cdots\times[a_d,b_d].
\]
Under the Gaussian corruption model $p(\obin\mid \nfin)=\mathcal N(\obin\mid \nfin,\Gamma)$,
the posterior is the truncated multivariate normal
\[
p(\nfin\mid \obin)
=
\frac{\mathbf 1_{\mathcal B}(\nfin)}{Z(\mathcal B;\obin,\Gamma)}
\,
\phi_d(\nfin;\obin,\Gamma),
\]
where
\[
\phi_d(\nfin;\obin,\Gamma)
=
\frac{1}{(2\pi)^{d/2}\det(\Gamma)^{1/2}}
\exp\!\Big(
-\tfrac12(\nfin-\obin)^\top\Gamma^{-1}(\nfin-\obin)
\Big)
\]
and
\[
Z(\mathcal B;\obin,\Gamma)
=
\int_{\mathcal B}\phi_d(u;\obin,\Gamma)\,du.
\]
\end{proposition}

\begin{proof}
The prior density is constant on $\mathcal B$ and zero outside, so Bayes' rule gives
\[
p(\nfin\mid \obin)
\propto
\mathbf 1_{\mathcal B}(\nfin)\,p(\obin\mid \nfin).
\]
Substituting the Gaussian corruption density yields
\[
p(\nfin\mid \obin)
\propto
\mathbf 1_{\mathcal B}(\nfin)\,\phi_d(\nfin;\obin,\Gamma).
\]
Normalizing over $\mathcal B$ gives the stated form.
\end{proof}

\section{ELBO Derivation}
\label{app:elbo}

We derive the evidence lower bound (ELBO) for \name{} in four steps:
first the full-data ELBO, then the context--target training objective.

\paragraph{Step 1: Marginal Likelihood.}
We aim to maximize the log-likelihood of the observed labels conditioned
on the noisy inputs. Starting from the joint model~\eqref{eq:dnp-joint},
we marginalize over the latent variables $z$ and $\nfin_{[n]}$:
\begin{align}
p(y_{[n]} \mid \obin_{[n]})
&= \iint p(z)\,p(\nfin_{[n]} \mid \obin_{[n]})
   \prod_{i=1}^n p(y_i \mid \nfin_i, z)\;dz\;d\nfin,
\label{eq:marginal}
\end{align}
where $p(\nfin_{[n]} \mid \obin_{[n]}) = p(\obin_{[n]} \mid \nfin_{[n]})\,
p(\nfin_{[n]})\,/\,p(\obin_{[n]})$ follows from Bayes' rule.

\paragraph{Step 2: Introducing the Variational Family.}
Since the integral in~\eqref{eq:marginal} is intractable, we introduce
the structured variational posterior from~\eqref{eq:var-family}
by multiplying and dividing inside the integral:
\begin{align}
\log p(y_{[n]} \mid \obin_{[n]})
&= \log \iint
\frac{q_\psi(\nfin_{[n]} \mid \obin_{[n]})\,
      q_\phi(z \mid \nfin_{[n]}, y_{[n]})}
     {q_\psi(\nfin_{[n]} \mid \obin_{[n]})\,
      q_\phi(z \mid \nfin_{[n]}, y_{[n]})}
\notag\\
&\qquad\quad \times\,
p(z)\,p(\nfin_{[n]} \mid \obin_{[n]})
\prod_{i=1}^n p(y_i \mid \nfin_i, z)\;dz\;d\nfin.
\end{align}
Recognizing the double integral as an expectation under
$q_\psi q_\phi$ and applying Jensen's inequality
($\log\mathbb{E}[\cdot]\ge\mathbb{E}[\log\cdot]$):
\begin{align}
\log p(y_{[n]} \mid \obin_{[n]})
&\ge
\mathbb{E}_{q_\psi q_\phi}
\Bigg[
  \log\frac{p(\nfin_{[n]} \mid \obin_{[n]})}
           {q_\psi(\nfin_{[n]} \mid \obin_{[n]})}
+ \log\frac{p(z)}
           {q_\phi(z \mid \nfin_{[n]}, y_{[n]})}
+ \sum_{i=1}^n \log p(y_i \mid \nfin_i, z)
\Bigg].
\label{eq:elbo-ratio}
\end{align}

\paragraph{Step 3: Rewriting as KL Divergences.}
The first two terms in~\eqref{eq:elbo-ratio} are negative KL
divergences by definition:
\begin{align}
\mathbb{E}_{q_\psi}\!\left[
  \log\frac{p(\nfin_{[n]}\mid\obin_{[n]})}
           {q_\psi(\nfin_{[n]}\mid\obin_{[n]})}
\right]
&= -\mathrm{KL}\!\Big(
     q_\psi(\nfin_{[n]}\mid\obin_{[n]})
     \;\|\;
     p(\nfin_{[n]}\mid\obin_{[n]})
   \Big),
\\
\mathbb{E}_{q_\phi}\!\left[
  \log\frac{p(z)}{q_\phi(z\mid \nfin_{[n]},y_{[n]})}
\right]
&= -\mathrm{KL}\!\Big(
     q_\phi(z\mid \nfin_{[n]},y_{[n]})
     \;\|\;
     p(z)
   \Big).
\end{align}
Note that $p(\nfin_{[n]}\mid\obin_{[n]})$ factors as
$\prod_i p(\nfin_i\mid\obin_i)$ under the assumed independence
structure, so the KL decomposes point-wise. Furthermore, expanding
$p(\nfin_{[n]}\mid\obin_{[n]}) \propto
p(\obin_{[n]}\mid \nfin_{[n]})\,p(\nfin_{[n]})$ and absorbing the
$\log p(\obin_{[n]}\mid \nfin_{[n]})$ term into the reconstruction
objective yields the form stated in~\eqref{eq:dnp-elbo-full}.

\paragraph{Step 4: Context--Target Split.}
For meta-learning we condition on the context set and predict targets.
Splitting indices into context $C=\{1,\dots,m\}$ and
target $T=\{m+1,\dots,n\}$, and replacing the prior $p(z)$ with
the context posterior $q_\phi(z\mid \nfin_C, y_C)$ as the baseline
(following the NP training objective~\eqref{eq:np-elbo}):
\begin{align}
\log p(y_T \mid \obin_{[n]}, y_C)
&\ge
\mathbb{E}_{q_\psi(\nfin_{[n]}\mid\obin_{[n]})}
\Bigg[
  \mathbb{E}_{q_\phi(z\mid \nfin_{[n]},y_{[n]})}
  \sum_{i\in T} \log p(y_i \mid \nfin_i, z)
\notag\\
&\qquad
  -\mathrm{KL}\!\Big(
    q_\phi(z \mid \nfin_{[n]}, y_{[n]})
    \;\|\;
    q_\phi(z \mid \nfin_C, y_C)
  \Big)
\Bigg]
\notag\\
&-\mathrm{KL}\!\Big(
    q_\psi(\nfin_{[n]}\mid\obin_{[n]})
    \;\|\;
    p(\nfin_{[n]})
  \Big)
+
\mathbb{E}_{q_\psi}
\sum_{i=1}^n \log p(\obin_i \mid \nfin_i),
\label{eq:elbo-ct-app}
\end{align}
which matches~\eqref{eq:dnp-elbo-ct} in the main text.

\section{Auxiliary Lemmas for Section~\ref{sec:theory}}
\label{app:theory-lemmas}
Here, we provide the lemmas that are used to prove the theorems and corollaries in Section~\ref{sec:theory}.
\begin{lemma}[Gaussian cross-entropy]
\label{lem:gaussian-cross-entropy}
Let $P$ be an absolutely continuous distribution on $\mathbb R$ with finite mean $\mu_P$ and variance $v_P$.
For $Q=\mathcal N(m,\sigma^2)$ with density $q$, one has
\begin{equation}
-\mathbb E_{Y\sim P}[\log q(Y)]
=
\frac12\log(2\pi\sigma^2)
+
\frac{v_P+(\mu_P-m)^2}{2\sigma^2}.
\label{eq:gaussian-cross-entropy}
\end{equation}
\end{lemma}

\begin{proof}
Since
\[
-\log q(Y)
=
\frac12\log(2\pi\sigma^2)
+
\frac{(Y-m)^2}{2\sigma^2},
\]
taking expectation under $P$ gives
\[
-\mathbb E_P[\log q(Y)]
=
\frac12\log(2\pi\sigma^2)
+
\frac{\mathbb E_P[(Y-m)^2]}{2\sigma^2}.
\]
Using the bias--variance decomposition
\[
\mathbb E_P[(Y-m)^2]
=
\mathrm{Var}_P(Y)+(\mathbb E_P[Y]-m)^2
=
v_P+(\mu_P-m)^2
\]
proves \eqref{eq:gaussian-cross-entropy}.
\end{proof}

\begin{lemma}[Entropy upper bound at fixed variance]
\label{lem:max-entropy}
Let $P$ be an absolutely continuous distribution on $\mathbb R$ with finite mean $\mu_P$ and finite variance $v_P$.
Then
\begin{equation}
H(P)\le \frac12\log(2\pi e\,v_P),
\label{eq:max-entropy}
\end{equation}
with equality if and only if $P$ is Gaussian with variance $v_P$.
\end{lemma}

\begin{proof}
Let $G=\mathcal N(\mu_P,v_P)$ with density $g$.
Since $\mathrm{KL}(P\|G)\ge 0$,
\[
-H(P)-\mathbb E_P[\log g(Y)]\ge 0.
\]
Applying Lemma~\ref{lem:gaussian-cross-entropy} with $m=\mu_P$ and $\sigma^2=v_P$ gives
\[
-\mathbb E_P[\log g(Y)]
=
\frac12\log(2\pi v_P)+\frac{v_P}{2v_P}
=
\frac12\log(2\pi e\,v_P).
\]
Hence
\[
H(P)\le \frac12\log(2\pi e\,v_P).
\]
Equality holds if and only if $\mathrm{KL}(P\|G)=0$, i.e.\ $P=G$ almost everywhere.
\end{proof}

\begin{lemma}[Third-order Taylor expansion with Lipschitz Hessian]
\label{lem:taylor}
Let $f:\mathbb R^d\to\mathbb R$ be twice continuously differentiable and assume its Hessian is globally $L$-Lipschitz:
\[
\|\nabla^2 f(u)-\nabla^2 f(v)\|\le L\|u-v\|
\qquad\text{for all }u,v\in\mathbb R^d.
\]
Fix $u_0\in\mathbb R^d$, and set $g=\nabla f(u_0)$ and $H=\nabla^2 f(u_0)$.
Then for every $h\in\mathbb R^d$,
\begin{equation}
f(u_0+h)
=
f(u_0)+g^\top h+\frac12 h^\top H h + r(h),
\label{eq:taylor-lemma}
\end{equation}
where
\begin{equation}
|r(h)|\le \frac{L}{6}\|h\|^3.
\label{eq:taylor-remainder}
\end{equation}
\end{lemma}

\begin{proof}
By the integral form of Taylor's theorem,
\[
f(u_0+h)
=
f(u_0)+g^\top h
+
\int_0^1 (1-t)\, h^\top \nabla^2 f(u_0+th)\, h\,dt.
\]
Subtracting $\frac12 h^\top Hh = \int_0^1 (1-t)h^\top Hh\,dt$ gives
\[
r(h)
=
\int_0^1 (1-t)\,h^\top\big(\nabla^2 f(u_0+th)-H\big)h\,dt.
\]
Hence
\[
|r(h)|
\le
\int_0^1 (1-t)\,\|\nabla^2 f(u_0+th)-H\|\,\|h\|^2\,dt
\le
\int_0^1 (1-t)\,Lt\,dt\,\|h\|^3
=
\frac{L}{6}\|h\|^3.
\]
\end{proof}

\begin{lemma}[Gaussian moment bounds]
\label{lem:gaussian-moments}
Let $\Delta\sim \mathcal N(0,S)$ in $\mathbb R^d$ with $S\succeq 0$.
For every integer $k\ge 1$,
\begin{equation}
\mathbb E\|\Delta\|_2^k
\le
C_{d,k}\,\|S\|^{k/2},
\label{eq:gaussian-moment-bound}
\end{equation}
where $C_{d,k}:=\mathbb E\|G\|_2^k<\infty$ for $G\sim \mathcal N(0,I_d)$.
\end{lemma}

\begin{proof}
Write $\Delta=S^{1/2}G$ with $G\sim \mathcal N(0,I_d)$.
Then
\[
\|\Delta\|_2
=
\|S^{1/2}G\|_2
\le
\|S^{1/2}\|\,\|G\|_2
=
\|S\|^{1/2}\|G\|_2.
\]
Raising both sides to the $k$th power and taking expectation yields
\[
\mathbb E\|\Delta\|_2^k
\le
\|S\|^{k/2}\,\mathbb E\|G\|_2^k
=
C_{d,k}\,\|S\|^{k/2}.
\]
\end{proof}

\begin{lemma}[Odd symmetry for centered Gaussians]
\label{lem:odd-symmetry}
Let $\Delta\sim \mathcal N(0,S)$ in $\mathbb R^d$.
If $h:\mathbb R^d\to\mathbb R$ is measurable, integrable, and odd, i.e.\ $h(-u)=-h(u)$ for all $u$, then
\begin{equation}
\mathbb E[h(\Delta)]=0.
\label{eq:odd-symmetry}
\end{equation}
In particular, for every $g\in\mathbb R^d$ and every symmetric $H\in\mathbb R^{d\times d}$,
\begin{equation}
\mathbb E\!\big[(g^\top\Delta)(\Delta^\top H\Delta)\big]=0.
\label{eq:odd-symmetry-quadratic}
\end{equation}
\end{lemma}

\begin{proof}
Since $\Delta$ is centered Gaussian, $\Delta \overset{d}{=} -\Delta$.
Therefore
\[
\mathbb E[h(\Delta)]
=
\mathbb E[h(-\Delta)]
=
-\mathbb E[h(\Delta)],
\]
which implies \eqref{eq:odd-symmetry}.
For \eqref{eq:odd-symmetry-quadratic}, note that
\[
u\mapsto (g^\top u)(u^\top H u)
\]
is odd because the linear factor changes sign and the quadratic factor does not.
\end{proof}

\section{Proofs for Section~\ref{sec:theory}}
\label{app:theory-proofs}
In this section, we prove our theorems and corollaries.
\subsection{Proof of Theorem~\ref{thm:realizability}}
\label{app:proof-realizability}

By the definition of the \name{} predictive class,
\[
q_{\theta^\star,\psi^\star}^{\mathrm{\name{}}}(y\mid \obin,z)
=
\int
\mathcal N\!\big(y\mid f_{\theta^\star}(\nfin,z),\sigma_y^2\big)\,
q_{\psi^\star}(\nfin\mid \obin)\,d\nfin.
\]
Using the realizability assumptions \eqref{eq:realizability-assumptions},
\[
q_{\theta^\star,\psi^\star}^{\mathrm{\name{}}}(y\mid \obin,z)
=
\int
\mathcal N\!\big(y\mid f^\star(\nfin,z),\sigma_y^2\big)\,
p(\nfin\mid \obin)\,d\nfin.
\]
The right-hand side is exactly the Bayes-optimal predictive conditional from Section~\ref{sec:theory}, proving \eqref{eq:realizability-pointwise}.
The KL divergence therefore vanishes pointwise, and taking expectation over $(\obin,Z)$ gives \eqref{eq:realizability-kl}.

\subsection{Proof of Theorem~\ref{thm:plugin-kl}}
\label{app:proof-plugin-kl}

Fix $(\obin,z)$ and abbreviate
\[
P:=P_{\obin,z},
\qquad
Q:=\mathcal N(m(\obin,z),\sigma_y^2),
\qquad
\mu_\star:=\mu_\star(\obin,z),
\qquad
v_\star:=v_\star(\obin,z),
\qquad
m:=m(\obin,z).
\]
By the Bayes-optimal predictive conditional, $P$ is a Gaussian mixture and is therefore absolutely continuous.
Let $q$ denote the density of $Q$.

By definition,
\[
\mathrm{KL}(P\|Q)
=
-H(P)-\mathbb E_{Y\sim P}[\log q(Y)].
\]
Applying Lemma~\ref{lem:gaussian-cross-entropy} gives
\[
-\mathbb E_{Y\sim P}[\log q(Y)]
=
\frac12\log(2\pi\sigma_y^2)
+
\frac{v_\star+(\mu_\star-m)^2}{2\sigma_y^2}.
\]
Applying Lemma~\ref{lem:max-entropy} to $P$ yields
\[
H(P)\le \frac12\log(2\pi e\,v_\star).
\]
Combining the last two displays,
\begin{align*}
\mathrm{KL}(P\|Q)
&\ge
-\frac12\log(2\pi e\,v_\star)
+
\frac12\log(2\pi\sigma_y^2)
+
\frac{v_\star+(\mu_\star-m)^2}{2\sigma_y^2}\\
&=
\frac{(\mu_\star-m)^2}{2\sigma_y^2}
+
\frac12
\left(
\frac{v_\star}{\sigma_y^2}
-1
-\log\frac{v_\star}{\sigma_y^2}
\right).
\end{align*}

The right-hand side is minimized over $m$ by choosing $m=\mu_\star$, proving \eqref{eq:kl-lb-varonly}.

It remains to verify the strict-positivity claim.
By \eqref{eq:bayes-var},
\[
v_\star
=
\sigma_y^2+\mathrm{Var}\!\big(f^\star(X,z)\mid \obin\big).
\]
If $\mathrm{Var}(f^\star(X,z)\mid \obin) > 0$, then $v_\star>\sigma_y^2$.
Define
\[
\phi(u):=u-1-\log u,\qquad u>0.
\]
Then
\[
\phi'(u)=1-\frac1u,\qquad \phi''(u)=\frac{1}{u^2}>0.
\]
Hence $\phi$ is strictly convex and attains its unique minimum $0$ at $u=1$.
Therefore $\phi(u)>0$ for every $u\neq 1$, and in particular
\[
\frac12
\left(
\frac{v_\star}{\sigma_y^2}
-1
-\log\frac{v_\star}{\sigma_y^2}
\right)
>0
\qquad\text{whenever } v_\star>\sigma_y^2.
\]
This completes the proof.

\subsection{Proof of Lemma~\ref{lem:moment-expansion}, Theorem~\ref{thm:scaling-kl}, and Corollary~\ref{cor:scaling-isotropic}}
\label{app:proof-scaling}
For brevity, write $\bar\nfin:=\mu_{\mathrm{post}}(\obin)$, $S:=\Sigma_{\mathrm{post}}$, and $f(\cdot):=f^\star(\cdot,z)$.
By Proposition~\ref{prop:gaussian-posterior},
\[
X\mid \obin \sim \mathcal N(\bar\nfin,S).
\]
Define
\[
\Delta:=X-\bar\nfin,
\qquad
g:=\nabla f(\bar\nfin),
\qquad
H:=\nabla^2 f(\bar\nfin).
\]
Then $\Delta\sim \mathcal N(0,S)$.

Since $\nabla^2 f$ is globally Lipschitz, Lemma~\ref{lem:taylor} gives
\begin{equation}
f(\bar\nfin+\Delta)
=
f(\bar\nfin)+g^\top\Delta+\frac12\Delta^\top H\Delta+r(\Delta),
\label{eq:scaling-taylor}
\end{equation}
with
\begin{equation}
|r(\Delta)|\le C\|\Delta\|^3
\label{eq:scaling-remainder}
\end{equation}
for a constant $C$ independent of $\Delta$.

\paragraph{Proof of \eqref{eq:scaling-mean}.} Taking expectation in \eqref{eq:scaling-taylor} yields
\[
\mathbb E[f(X)\mid \obin]
=
f(\bar\nfin)
+
\mathbb E[g^\top\Delta]
+
\frac12\mathbb E[\Delta^\top H\Delta]
+
\mathbb E[r(\Delta)].
\]
Because $\Delta$ is centered Gaussian,
\[
\mathbb E[g^\top\Delta]=0,
\qquad
\mathbb E[\Delta^\top H\Delta]=\mathrm{tr}(HS).
\]
Moreover, by \eqref{eq:scaling-remainder} and Lemma~\ref{lem:gaussian-moments},
\[
|\mathbb E[r(\Delta)]|
\le
C\,\mathbb E\|\Delta\|^3
=
O(\|S\|^{3/2}).
\]
Therefore
\[
\mathbb E[f(X)\mid \obin]
=
f(\bar\nfin)+\frac12\mathrm{tr}(HS)+O(\|S\|^{3/2}),
\]
which proves \eqref{eq:scaling-mean}.

\paragraph{Proof of \eqref{eq:scaling-var}.} Define the centered quadratic and remainder terms
\[
Q(\Delta):=
\frac12\Big(\Delta^\top H\Delta-\mathrm{tr}(HS)\Big),
\qquad
R(\Delta):=
r(\Delta)-\mathbb E[r(\Delta)].
\]
Then
\[
f(X)-\mathbb E[f(X)\mid \obin]
=
g^\top\Delta+Q(\Delta)+R(\Delta).
\]
Hence
\begin{align}
\mathrm{Var}(f(X)\mid \obin)
&=
\mathbb E\big[(g^\top\Delta)^2\big]
+
\mathbb E\big[Q(\Delta)^2\big]
+
\mathbb E\big[R(\Delta)^2\big]
\notag\\
&\quad
+
2\mathbb E\big[g^\top\Delta\,Q(\Delta)\big]
+
2\mathbb E\big[g^\top\Delta\,R(\Delta)\big]
+
2\mathbb E\big[Q(\Delta)R(\Delta)\big].
\label{eq:scaling-var-decomp}
\end{align}
We now bound each term.

First,
\[
\mathbb E[(g^\top\Delta)^2]=g^\top S g.
\]

Second, since
\[
|Q(\Delta)|
\le
\frac12|\Delta^\top H\Delta|
+
\frac12|\mathrm{tr}(HS)|
\le
\frac12\|H\|\,\|\Delta\|^2
+
\frac12|\mathrm{tr}(HS)|,
\]
we have
\[
\mathbb E[Q(\Delta)^2]
=
O\big(\mathbb E\|\Delta\|^4\big)+O(\|S\|^2)
=
O(\|S\|^2),
\]
where Lemma~\ref{lem:gaussian-moments} was used in the last step.

Third, by \eqref{eq:scaling-remainder},
\[
|R(\Delta)|
\le
|r(\Delta)|+|\mathbb E[r(\Delta)]|
\le
C\|\Delta\|^3 + C\mathbb E\|\Delta\|^3.
\]
Therefore
\[
\mathbb E[R(\Delta)^2]
=
O\big(\mathbb E\|\Delta\|^6\big)
=
O(\|S\|^3)
\]
by Lemma~\ref{lem:gaussian-moments}.

Fourth, by Lemma~\ref{lem:odd-symmetry},
\[
\mathbb E[g^\top\Delta\,Q(\Delta)]
=
\frac12\mathbb E\big[(g^\top\Delta)(\Delta^\top H\Delta)\big]
-
\frac12\mathrm{tr}(HS)\,\mathbb E[g^\top\Delta]
=
0.
\]

Fifth, by Cauchy--Schwarz,
\[
\big|\mathbb E[g^\top\Delta\,R(\Delta)]\big|
\le
\Big(\mathbb E[(g^\top\Delta)^2]\Big)^{1/2}
\Big(\mathbb E[R(\Delta)^2]\Big)^{1/2}
=
O(\|S\|^{1/2})\,O(\|S\|^{3/2})
=
O(\|S\|^2).
\]

Sixth, again by Cauchy--Schwarz,
\[
\big|\mathbb E[Q(\Delta)R(\Delta)]\big|
\le
\Big(\mathbb E[Q(\Delta)^2]\Big)^{1/2}
\Big(\mathbb E[R(\Delta)^2]\Big)^{1/2}
=
O(\|S\|)\,O(\|S\|^{3/2})
=
O(\|S\|^{5/2}).
\]

Substituting these bounds into \eqref{eq:scaling-var-decomp} gives
\[
\mathrm{Var}(f(X)\mid \obin)
=
g^\top S g + O(\|S\|^2),
\]
which proves \eqref{eq:scaling-var}.

\paragraph{Proof of \eqref{eq:scaling-kl}.}
Let $m_0:=f(\bar\nfin)$.
By Theorem~\ref{thm:plugin-kl},
\begin{equation}
\mathrm{KL}\!\big(P_{\obin,z}\ \|\ \mathcal N(m_0,\sigma_y^2)\big)
\ge
\frac{(\mu_\star-m_0)^2}{2\sigma_y^2}
+
\frac12\left(
\frac{v_\star}{\sigma_y^2}
-1
-\log\frac{v_\star}{\sigma_y^2}
\right).
\label{eq:scaling-kl-start}
\end{equation}
By \eqref{eq:scaling-mean},
\[
\mu_\star-m_0
=
\frac12\mathrm{tr}(HS)+O(\|S\|^{3/2}),
\]
so
\begin{equation}
\frac{(\mu_\star-m_0)^2}{2\sigma_y^2}
=
\frac{1}{8\sigma_y^2}\big(\mathrm{tr}(HS)\big)^2 + o(\|S\|^2).
\label{eq:scaling-kl-bias}
\end{equation}

Next, let $a:=v_\star-\sigma_y^2$.
By \eqref{eq:scaling-var},
\[
a=g^\top S g + O(\|S\|^2),
\qquad
a=O(\|S\|).
\]
Write $u:=a/\sigma_y^2$.
Then $u=O(\|S\|)$, and
\[
\frac12\left(
\frac{v_\star}{\sigma_y^2}
-1
-\log\frac{v_\star}{\sigma_y^2}
\right)
=
\frac12\big(u-\log(1+u)\big).
\]
Since $u\to 0$, Taylor expansion of $\log(1+u)$ at $0$ gives
\[
u-\log(1+u)=\frac{u^2}{2}+O(u^3),
\]
hence
\begin{equation}
\frac12\left(
\frac{v_\star}{\sigma_y^2}
-1
-\log\frac{v_\star}{\sigma_y^2}
\right)
=
\frac{1}{4\sigma_y^4}(g^\top S g)^2 + o(\|S\|^2).
\label{eq:scaling-kl-varterm}
\end{equation}
Combining \eqref{eq:scaling-kl-start}, \eqref{eq:scaling-kl-bias}, and \eqref{eq:scaling-kl-varterm} proves \eqref{eq:scaling-kl}.

\paragraph{Proof of \eqref{eq:scaling-kl-irreducible} and Corollary~\ref{cor:scaling-isotropic}.}
\eqref{eq:scaling-kl-irreducible} follows directly from \eqref{eq:kl-lb-varonly} and the expansion \eqref{eq:scaling-kl-varterm}.
If $S=\tau^2 I_d$, then
\[
g^\top S g=\tau^2\|g\|_2^2,
\]
so the isotropic scaling stated in Corollary~\ref{cor:scaling-isotropic} follows immediately.

\section{Experimental Detail}
\label{app:experiment}
All experiments were run on a single NVIDIA H200 GPU on an internal computing cluster. The training and inference time of \name{} is approximately the same as that of standard NPs, since the additional denoiser is a small MLP applied per input and contributes negligible overhead relative to the encoder and decoder networks.

\begin{table}[t]
\caption{Overview of datasets used in experiments. $d$ denotes the input dimensionality, $K$ the number of training tasks (episodes), Task specifies the type of problem (regression, image completion, or spatial interpolation), and Noise indicates whether inputs are naturally corrupted. MNIST and 1D GP datasets do not inherently contain input noise, so we add noise synthetically to simulate noisy measurements.}
\begin{center}
\begin{small}
\begin{sc}
\begin{tabular}{lccccc}
\toprule
Dataset & $K$ & $d$ & Task & Noise \\
\midrule
1D GP       & 10,000 & 1  & Regression       & No  \\
MNIST            & 70,000 & 2  & Image Compl.     & No  \\
Air Quality      & 2,856  & 4  & Spatial Interp.  & Yes \\
NOAA CA 2026     & 90     & 4  & Spatial Interp.  & Yes \\
\bottomrule
\end{tabular}
\label{tbl:dataset}
\end{sc}
\end{small}
\end{center}
\vskip -0.1in
\end{table}

\newpage

\end{document}